\documentclass[sigconf]{acmart}

\usepackage{xcolor}
\usepackage[ruled]{algorithm2e}
\usepackage{algpseudocode}
\usepackage{multirow}
\usepackage{multicol}
\usepackage{enumitem}

\AtBeginDocument{%
  \providecommand\BibTeX{{%
    \normalfont B\kern-0.5em{\scshape i\kern-0.25em b}\kern-0.8em\TeX}}}

\setcopyright{acmcopyright}
\copyrightyear{2022}
\acmYear{2022}
\setcopyright{acmlicensed}\acmConference[KDD '22]{Proceedings of the 28th ACM SIGKDD Conference on Knowledge Discovery and Data Mining}{August 14--18, 2022}{Washington, DC, USA}
\acmBooktitle{Proceedings of the 28th ACM SIGKDD Conference on Knowledge Discovery and Data Mining (KDD '22), August 14--18, 2022, Washington, DC, USA}
\acmPrice{15.00}
\acmDOI{10.1145/3534678.3539175}
\acmISBN{978-1-4503-9385-0/22/08}

\begin{document}

\title{Interpretable Personalized Experimentation}

\author{Han Wu$^\dagger$}
\authornote{ Work conducted during an internship at Meta.\\
$^\dagger$ Equal contribution.\\
$^\ddagger$ Work done while employed by Meta.}
\email{hanwu71@stanford.edu}
\affiliation{
\institution{Stanford University}
\city{Stanford}
\state{CA}
\country{USA}
}

%\author{Sarah Tan$^\dagger$, Weiwei Li, Mia Garrard, Adam Obeng$^\ddagger$, Drew Dimmery$^\ddagger$, Shaun Singh$^\ddagger$, Hanson Wang$^\ddagger$, Daniel Jiang, Eytan Bakshy}
%\affiliation{
%\institution{Meta}
%\city{Menlo Park}
%\state{CA}
%\country{USA}
%}

\author{Sarah Tan$^\dagger$}
\email{sarahtan@fb.com}
\affiliation{
\institution{Meta}
\city{Menlo Park}
\state{CA}
\country{USA}
}

\author{Weiwei Li}
\email{weiweili90@fb.com}
\affiliation{
\institution{Meta}
\city{Menlo Park}
\state{CA}
\country{USA}
}

\author{Mia Garrard}
\email{mgarrard@fb.com}
\affiliation{
\institution{Meta}
\city{Menlo Park}
\state{CA}
\country{USA}
}

\author{Adam Obeng$^\ddagger$ }
\email{adam@adamobeng.com}
\affiliation{
\institution{}
\city{San Francisco}
\state{CA}
\country{USA}
}

\author{Drew Dimmery$^\ddagger$ }
\email{drew.dimmery@gmail.com}
\affiliation{
\institution{University of Vienna}
\city{Vienna}
\state{}
\country{Austria}
}

\author{Shaun Singh$^\ddagger$ }
\email{shaundsingh@gmail.com}
\affiliation{
\institution{}
\city{San Francisco}
\state{CA}
\country{USA}
}

\author{Hanson Wang$^\ddagger$ }
\email{hanson.wng@gmail.com}
\affiliation{
\institution{}
\city{San Francisco}
\state{CA}
\country{USA}
}

\author{Daniel Jiang}
\email{drjiang@fb.com}
\affiliation{
\institution{Meta}
\city{Menlo Park}
\state{CA}
\country{USA}
}

\author{Eytan Bakshy}
\email{ebakshy@fb.com}
\affiliation{
\institution{Meta}
\city{Menlo Park}
\state{CA}
\country{USA}
}

\renewcommand{\shortauthors}{Han Wu et al.}

\begin{abstract}
Black-box heterogeneous treatment effect (HTE) models are increasingly being used to create personalized policies that assign individuals to their optimal treatments. However, they are difficult to understand, and can be burdensome to maintain in a production environment. In this paper, we present a scalable, interpretable personalized experimentation system, implemented and deployed in production at Meta. The system works in a multiple treatment, multiple outcome setting typical at Meta to: (1) learn explanations for black-box HTE models; (2) generate interpretable personalized policies. We evaluate the methods used in the system on publicly available data and Meta use cases, and discuss lessons learnt during the development of the system. 

\end{abstract}

\newcommand{\note}[1]{[{\color{red}{Note:}} {\color{blue}{#1}}]}


%\ccsdesc[100]{Computing methodologies~Classification and regression trees}


\maketitle

\vspace{-0.1cm}
\section{Introduction}
\label{sec: introduction}

In conventional A/B testing, individuals are randomly assigned to treatment groups, with the goal of selecting a single best treatment (on average) for the entire population. Increasingly, internet companies are taking a \emph{personalized} approach, making use of heterogeneous treatment effects models (HTE) that predict individual-level treatment effects for each individual, and personalized policies that aim to deliver the best treatment to each individual \citep{garcin2013personalized, grbovic2018real}. 

Current state-of-the-art HTE models used for personalization frequently leverage black-box, un-interpretable base learners such as gradient boosted trees and neural networks \citep{wager2018estimation, kennedy2020optimal}. Some systems even generate HTE predictions by combining outputs from \emph{multiple} models via ensembling or meta-learning (e.g., stacking, weighting) \citep{kunzel2019metalearners, nie2021quasi, montoya2021optimal}. Moreover, the complete end-to-end personalization system (from individual features to treatment decisions) sometimes uses these treatment effect predictions as inputs to additional black-box policy learning models \citep{imai2011estimation}. 

\begin{figure*}
  \begin{center}
    \includegraphics[width=1\textwidth]{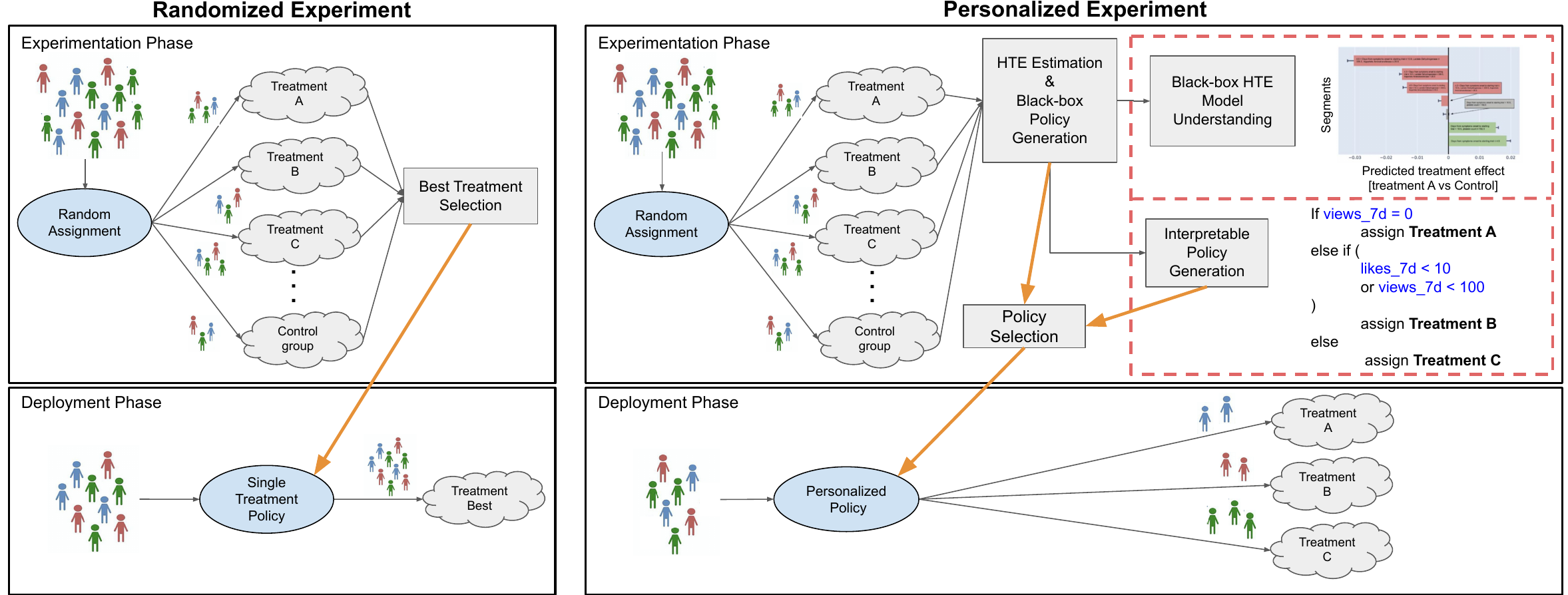}
  \end{center}
  \caption{\emph{Left}: Standard randomized experiment. During the experimentation phase, individuals are randomly assigned to treatments and control, and the best treatment is selected. At the  deployment phase, that treatment is applied to all individuals. \emph{Right}: Personalized experiment. At experimentation time we generate a policy that will decide, at deployment time, what treatment to use for each individual. Our contributions are depicted in the dashed red box: i) an approach to understand existing black-box HTE policies, to empower product teams to decide whether they can be trusted, and ii) an approach to generate interpretable policies, that can be used instead of the black-box policies.}
 \label{fig:main_diagram}
\end{figure*}

There are several challenges with deploying these black-box approaches at internet companies. First, black-box HTE models and resulting policies are difficult to interpret, which can deter their uptake. Second, as dataset size and the number of features increases, which is common in internet companies, the burden of maintaining such models in production increases \citep{sculley2015hidden}. Although there exists work that addresses these challenges via optimal interpretable policy learning methods \citep{amram2020optimal, zhou2018offline}, we have found these methods computationally intractable on large-scale datasets. 

In this paper we present a scalable, interpretable personalized experimentation system, implemented and deployed in production at Meta. The system is divided into a HTE model interpretation stage and an interpretable policy generation stage. Our HTE model interpretation approach is responsible for helping product teams understand how black-box HTE models are performing personalization (deciding which treatment group to assign to each user). After that, our interpretable policy generation approach generates segments of individuals to recommend different treatment groups to, to capture potential gains from personalization. A product team then has the choice of deploying a black-box policy (implied by the black-box HTE model) or an interpretable policy generated by our methods. Figure 1 shows the system.

%The rest of the paper is organized as follows: Sec. 2 overviews related work. Sec. 3 describes the problem setting of large-scale experimentation. Sec. 4 describes our HTE model interpretation and interpretable policy generation system. Sec. 5 describes experimental evaluation on publicly available data. Sec. 6 describes system deployment at Meta, and lessons learnt. Sec. 7 concludes.

\section{Related Work}
\label{sec: related_work}
\textbf{Large-scale HTE personalization.} Many different types of HTE models have been proposed (see previous section), and different aspects of utilizing them for personalization in industry have been considered. For example, Lada et al. combined experimental and observational data to estimate individuals' heterogeneous response to page recommendations to personalize Facebook News Feed \cite{lada2019observational}. Xie et al. tackled the problem of false discovery, the chances of which increases with large-scale data \cite{xie2018false}. The focus of this paper is on increasing understanding of black-box HTE models, an area that has received less attention, and deployment of interpretable policies in lieu of black-box policies.

\textbf{Segment discovery.} Our work surfacing segments with HTE produces a representation analogous to segment discovery methods. These methods are divided into: (1) statistical tests to confirm or deny a set of pre-defined hypotheses about segments \citep{assmann2000subgroup, song2007method}; (2) methods that aim to discover segments directly from data \citep{imai2013estimating, chen2015prim, wang2017causal, nagpal2020interpretable}; (3) methods that act on predicted individual-level treatment effects to discover segments \cite{foster2011subgroup, lee2020robust, dwivedi2020stable}. However, the comparison of segment finding algorithms is still an open question. Loh et al. empirically compared 13 different segment-finding algorithms \citep{loh2019subgroup}, finding that no one algorithm satisfied all the desired properties of such algorithms. Closest to Distill-HTE, the method we use in our system, are \cite{foster2011subgroup, lee2020robust}; we compare these three methods in Sec. \ref{subsec: subgroups}.

\textbf{Policy learning.} Many methods have been proposed to learn policies from experimental or observational data. However, most methods produce black-box policies \citep{qian2011performance,zhao2012estimating,swaminathan2015counterfactual,kitagawa2018should,kallus2018policy}. Several existing works \citep{amram2020optimal, zhou2018offline, athey2021policy, jo2021learning} construct interpretable policies with similar representations (segments) as the methods we use in this paper, but do so by constructing exact optimal trees, which is NP-hard \citep{laurent1976constructing}. Other works relax this by constructing approximate solutions to the exact optimal trees \cite{bertsimas2019optimal, dudiklangford2011}. In practice, we found these exact or approximately exact methods prohibitively slow on large-scale data and unsuitable for a production environment (see Section \ref{subsec: scalability} for runtime results when we tried these methods at Meta). With a focus on scalability, in this paper we consider methods using greedy heuristics to approximate optimal trees, similar to \cite{bertsimas2019optimal,dudiklangford2011, Hazan2010}, methods using distillation techniques to approximate black-box HTE models, similar to \cite{biggs2021model, makar2019distillation}, and methods that directly learn policies from data, similar to \cite{kallus2017recursive}. We do not propose novel policy learning methods in this paper, instead focusing on describing the system we built that evaluates different methods on Meta use cases. 

\begin{figure*}
\centering
\includegraphics[width=0.85\linewidth]{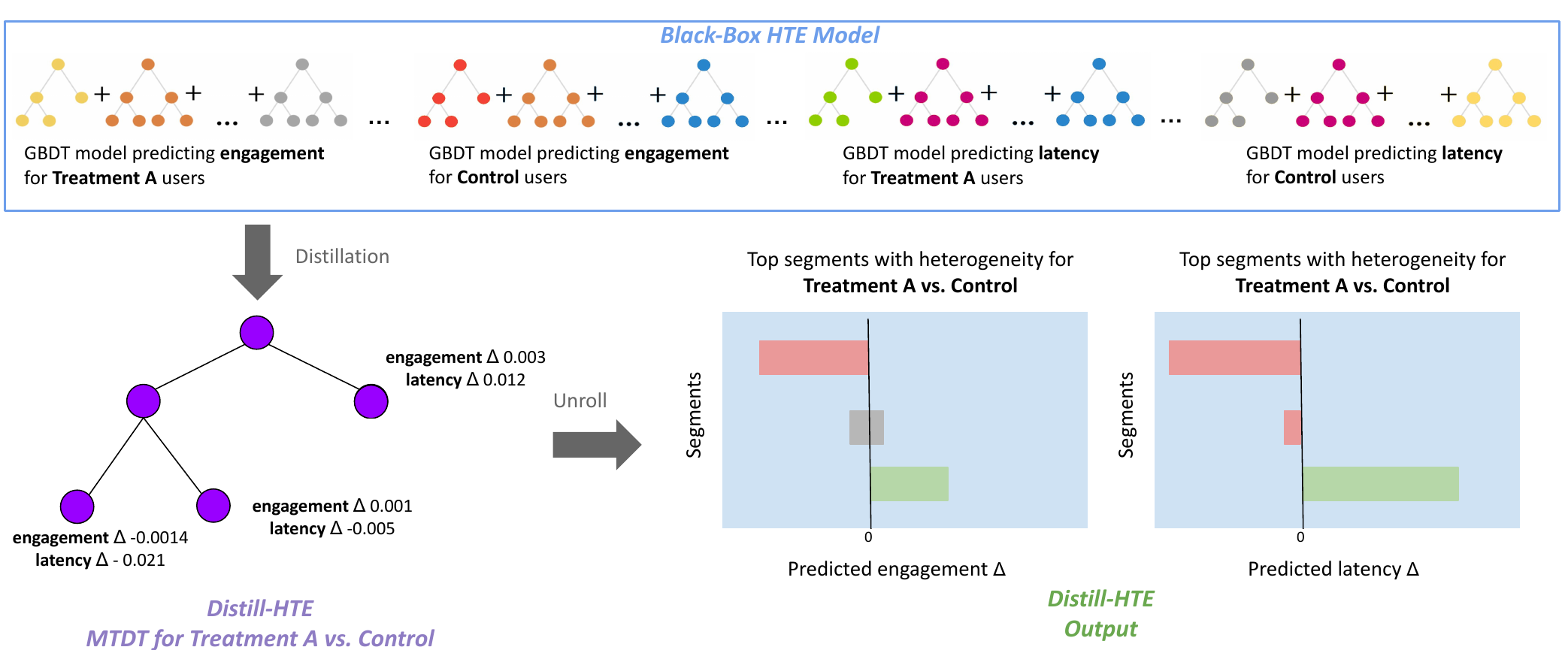}
\caption{Distill-HTE: Multitask decision tree (MTDT) learned by distilling treatment effect predictions from a black-box HTE model. After an MTDT is learned (\textit{bottom left}), we ``unroll'' the tree such that each terminal node is a bar (segment) in the barplot (\textit{bottom right}). Red and green bars are respectively segments with negative or positive predicted treatment effects; gray segments are not statistically significant. In our system, feature splits defining each segment (e.g. clicks$\_$7d $>$ 10 and views$\_$7d $>$ 100) are overlaid on top of each red/gray/green bar and also appear when a product team user hovers over the bar.}
\label{fig:MTDT}
\end{figure*}

\section{Problem Setting and Notation}
Our setting is that of \textbf{large-scale randomized experiments}, also called A/B tests, commonly conducted at internet companies to decide if a new product feature demonstrates gains over the existing product feature. An experiment is conducted by randomly assigning individuals to either the new product feature (treatment group A) or existing product feature (treatment group B), and computing changes in outcome values. It is common in our setting to have \textbf{more than two treatment groups and more than one outcome}. For example, a new product feature may increase engagement (one outcome) but use more resources (another outcome), so launching a policy requires weighing trade-offs between these outcomes. 

Concretely, denote the dataset collected from the experiment as $\mathcal{D}$, consisting of $N$ individuals randomly assigned to one of $K$ treatment groups or a control group, and $J$ outcomes of interest. Let $X_i\in \mathbb{R}^P$ be $P$ features of individual $i$, $W_i \in \{0,1,\ldots,K\}$ be the individual's treatment group assignment ($W_i = 0$ means control group), and $(Y_{i1},\ldots,Y_{iJ})$ be the individual's $J$ observed outcomes. Following the potential outcomes framework \citep{neyman1923applications,rubin1974estimating}, we assume for each individual $i$ and outcome $j$ that there are $K+1$ potential outcomes, $(Y_{ij}^{(0)},\ldots,Y_{ij}^{(K)})$. However, for each individual, we only observe one set of potential outcomes -- that of the treatment group that the individual was assigned to -- out of these $K+1$ potential outcomes. In other words, $(Y_{i1},\ldots,Y_{iJ})$ = $(Y_{i1}^{(W_i)},\ldots,Y_{iJ}^{(W_i)})$.

\textbf{Average Treatment Effects (ATE).} Let $\mathcal{G}_k$ denote the set of individuals in the $k$-th treatment group, $k = 0,1,\ldots, K$. The ATE of treatment $k$ compared to control on outcome $j$, computed using observed outcomes $Y_{ij}^{(k)}$, is: 
\label{eqn:ate}
\begin{equation} A_j^{(k)} = \displaystyle \sum_{i=1; i \in \mathcal{G}_k}^N \frac{Y_{ij}^{(k)}}{|\mathcal{G}_k|} - \sum_{i=1; i \in \mathcal{G}_{0}}^N \frac{Y_{ij}^{(0)}}{|\mathcal{G}_{0}|}\end{equation}

 \textbf{Heterogeneous Treatment Effects (HTE).} Product teams are not solely interested in average treatment effects, but also HTE, i.e. effects of new product features on particular groupings of individuals that are significantly different from the average individual. Reasons include ensuring that new product features are not showing gains on certain individuals at the expense of other individuals, delivering the best product feature to each individual, etc. 
For each individual $i$, outcome $j$, and treatment group $k$ compared to control, individual-level treatment effects are defined as
$\label{eqn:ite}
{T}_{ij}^{(k)} = {Y}_{ij}^{(k)} - {Y}_{ij}^{(0)}$ where ${T}_{ij}^{(0)}=0$. T-Learners \citep{athey2016recursive,kunzel2019metalearners}, a type of HTE model, replace ${Y}_{ij}^{(k)}$ by $\hat{Y}_{ij}^{(k)}$, the plugin estimators of individual potential outcome surfaces. Throughout this paper, we use $\hat{T}_{ij}$ to refer to any black-box estimate of treatment effects, and $\hat{Y}_{ij}$ to refer to any black-box estimate of potential outcomes. Our system is agnostic to the specific choice of HTE model, as long as it outputs $\hat{T}_{ij}$ and $\hat{Y}_{ij}$.

Certain HTE models produce predictions not at the individual-level, but rather at the segment-level, i.e. a grouping of individuals. Denote a segment, $S$, to be a set of indices representing a group of individuals. 
In our system, the segments learnt do not overlap (i.e. each individual belongs to exactly one segment). 

\textbf{Deterministic policies.} Policy $\Pi:\mathbb{R}^{P} \to \lbrace 0,1,..,K \rbrace$ decides, for individual $i$ with features $X_i$, which of the $K+1$ treatment groups to assign to the individual. 
In our system, we learn deterministic policies rather than random policies to ensure a more consistent UX experience for individuals. 
We evaluate different policies using \textbf{offline policy evaluation} (OPE). Concretely, let $\mathcal{G}_{\Pi}$ denote the set of individuals in the dataset whose treatment group matches that prescribed by policy $\Pi$. We compute:
\begin{equation}
\label{eqn:ope} 
\frac{1}{N}\displaystyle\sum_{i=1; i \in \mathcal{G}_{\Pi}}^N\frac{Y_i}{p_i},
\end{equation}
where $p_i$ is the propensity score for individual $i$ ($p_i = \frac{1}{K+1}$ in a randomized experiment with $K+1$ groups).

\section{Interpretable Personalized Experimentation}
Our system performs interpretable personalized experimentation in two independent stages: black-box HTE model understanding, and interpretable policy generation. We now describe the two stages.

\subsection{Understanding Black-Box Heterogeneous Treatment Effect Models}
\label{sec:interpreter} 
\subsubsection{Considerations when designing the system} 
In developing a black-box HTE model understanding method for Meta, we had several considerations: \begin{itemize}[leftmargin=*]\item The method should be a post-hoc method that can be applied to any type of HTE model, as the HTE models may be refreshed and retrained frequently. \item The method should handle multiple treatment groups and multiple outcomes, as is typical in industry settings. \item Since the outputs of the method will be placed in front of product teams to aid their understanding of the black-box HTE models, they should be understandable by engineers, data scientists, and product managers, with little training. 
\end{itemize}

\subsubsection{Method}
\label{sec: interpreter_method} We use \textbf{multi-task decision trees} (MTDT), which extend single-task decision trees by combining the prediction loss on each outcome, across multiple outcomes, into a single scalar loss function. This simple representation is effective and suitable for locating segments where individuals have heterogeneity across multiple outcomes. In the MTDT, each task (label) is the predicted treatment effect for an outcome, and each node is a segment identified to have elevated or depressed treatment effects across multiple outcomes. Since we learn these trees using distillation techniques, we call the method \textbf{Distill-HTE}. Figure \ref{fig:MTDT} illustrates the method. 

\textbf{Learning objective.} We suppose we have access to pairwise predicted treatment effects $\hat{T}_{ij}$ from the black-box HTE model. We will learn one MTDT model for each treatment group compared to control. To learn an MTDT model in a HTE model-agnostic fashion, we leverage model distillation techniques, taking the HTE model as teacher and the MTDT model as student \citep{bucilua2006compression, hinton2014distilling}. Let $\mathbf{\hat{F}}$ be the prediction function of an MTDT model for treatment group $k$ compared to control, using the following distillation training objective that minimizes the loss between predicted treatment effects $\hat{T}_{ij}$ and $\mathbf{\hat{F}}$:
\begin{equation}\label{eqn:distillation_loss}
L =  \frac{1}{N} \sum_{i=1}^N \;  \; \sum_{j=1}^J c_j \cdot \textit{l}(\mathbf{\hat{F}}_{[j]}(X_i), \hat{T}_{ij}^{(k)})
\end{equation}
$\mathbf{\hat{F}}_{[j]}$ is the MTDT model's prediction for  outcome $j$. $c_j$ is a weight that encodes how much outcome $j$ contributes to the overall loss, and \textit{l} is the loss function used to train model $\hat{F}$ (mean squared error in this case). When $c_j = 1$, as our system assumes by default, the loss of each outcome contributes equally to the overall distillation loss. However, all outcomes still affect jointly the training of the model in a way that training one model for each outcome independently would not \citep{caruana1997multitask}.

\textbf{Improving robustness.} We do the following: (1) Terminal nodes without sufficient overlap (i.e. enough treatment AND control points) are post-pruned from the tree, and predictions are regenerated; (2) Confidence intervals are provided for treatment effects within each node; (3) Honesty criterion \citep{athey2016recursive}: splits and predictions are determined on different data splits. 

\textbf{Interpretable output.} An MTDT can be visualized as a tree with each node having $J$ predicted treatment effects, one per outcome, and edges being feature splits (Figure 2 bottom left). We interviewed product teams at Meta to determine the best way to present this (see Section \ref{sec: user_studies} for interview findings), which led us to design the barplot-style output (Figure 2 bottom right) for product teams. 

\subsection{From Treatment Effects to Policies} 
\label{sec:weights}
When there are only two treatment groups and one outcome, it is easy to derive a policy from treatment effect predictions. For example, a policy derived from one of the two barplots in Fig. \ref{fig:MTDT} bottom right could be:
\textit{For the red segment where predicted treatment effect on engagement is negative for Treatment A vs. Control, assign to Control. For the green segment, assign to Treatment A.} However, it is not as easy to derive policies from treatment effect predictions when there are $>2$ treatment groups or $>1$ outcome.

\textbf{Multiple outcomes.} If another outcome is added (e.g. latency in Figure 2), both outcomes have to be considered together to derive a policy. One way to tradeoff between multiple outcomes is to modify the weights $(c_1, \ldots, c_J)$ used. This technique, from the multi-objective optimization literature \cite{gunantara2018review} and commonly used for ranking models in industry \cite{covington2016youtube, freno2017zalando, TuBasu2021, tang2021value}, weighs and adds multiple outcomes to form a single outcome: $Y_i^{(W_i)} = c_1 Y_{i1}^{(W_i)} + \cdots + c_J Y_{iJ}^{(W_i)}$

In the second stage of our system, we combine predicted multiple outcomes from HTE models this way to generate policies. The weights $(c_1, \ldots, c_J)$ can be the same weights used when training MTDT (Equation \eqref{eqn:distillation_loss}), or provided by product teams, reflecting their preferences for tradeoffs (e.g. no more than $x$ loss in latency to achieve $y$ gain in the engagement). They could also be assumed to be 1 by default (equally weighted), or tuned online \citep{letham2019bayesian}.

\textbf{Multiple treatment groups.} $K+1$ treatment groups and $J$ outcomes yields $K$ MTDT models, with each model predicting $J$ outcome treatment effects for each treatment group compared to control. Generating a single policy would require combining multiple MTDT, and it is not obvious the best way to do. Hence, while sufficient for understanding treatment effect predictions, MTDT is not suited for policy generation when there are more than two treatment groups. This motivates the second stage of our system: directly learning \textit{a single policy} that can assign more than two treatment groups.

\subsection{Interpretable Policy Generation}
\label{sec: interpretable_policies}
\subsubsection{Considerations when designing the system} 
\label{subsec: scalability}
In developing interpretable policy generation methods for Meta, we had several considerations:
\begin{itemize}[leftmargin=*]
\item Like the HTE model understanding method described in the previous section, the interpretable policy generation method should handle multiple treatment groups and multiple outcomes, and moreover, do so in one single model (the learned policy). Achieving this allows us to deploy only one policy in production, rather than multiple policies and then having to reconcile between them, which adds tech debt.
\item Unlike the HTE model understanding method which is run immediately after the black-box HTE model is trained, product teams would like to repeatedly interact with and tune the number of segments, size of each segment, weights to trade off multiple outcomes, etc. 
\item The method must be able to handle large-scale datasets with millions of individuals and hundreds of features.
\end{itemize}

We first considered optimal tree-based policy learning methods, as reviewed in Section \ref{sec: related_work}. However, in our initial experiments on a small sample of 100 thousand data points with 12 features and three treatment groups, \citep{zhou2018offline} that solves an exact tree search problem took 2.5 hours to run. With a quadratic runtime in the number of data points, it was impossible to run on our datasets that sometimes exceed 100 million data points. Other methods that find approximate solutions to exact optimal trees using coordinate ascent should be faster but their software is proprietary \citep{amram2020optimal, bertsimas2019optimal}. 

In the end, we implemented different methods in our system: (1) a method that approximates optimal tree finding methods with greedy heuristics (GreedyTreeSearch-HTE); (2) a method using distillation techniques to approximate black-box HTE models (Distill-Policy); (3) a method
that directly learns policies from data (No-HTE); (4) a method that ensembles different interpretable policies while remaining interpretable (GUIDE). All methods produced personalized policies that we visualize as a sequence of rules (Figure 3 top). All of these methods were inspired by close counterparts in literature, but we implemented them in slightly different ways to better suit our setting. We review this in detail in the next section. 

\subsubsection{Methods that take HTE model predictions as input} \label{sec:model_based_policy} The methods proposed in this section need an already trained HTE model from which predicted outcomes $\hat{Y}_{i}$ can be obtained.

\textit{\textbf{GreedyTreeSearch-HTE}} directly solves the following optimization problem in the space of trees with pre-determined maximum tree depth: $
    \Pi^* = \textnormal{argmax}_{\Pi} \frac{1}{N}\sum_{i=1}^{N} \hat{Y}_i^{(\Pi(X_i))}.
$
assuming without loss of generality that higher outcomes are better.
This can be treated as a cost-sensitive classification problem (NP-hard), where the cost of assigning a point to a group is the negative value of the predicted outcome \citep{foundation_cost_sensitive}. To achieve a scalable implementation, we solve the optimization problem greedily instead of resorting to exact tree search over the whole space of possible trees, and obtain personalized policy $\Pi$. See Algorithm~\ref{algo:model_distill_prediction} in the Appendix for the implementation. Other methods have been proposed to approximately solve this problem, for example those used in \citep{dudiklangford2011}. However, the greedy heuristic we chose offers the most simplicity and scalability.
 
\textit{\textbf{Distill-Policy}} starts from the naive policy $\Pi$ implied by the outcome predictions $\hat{Y}_i^{(0)},\ldots,\hat{Y}_i^{(K)}$, i.e. $\Pi_{HTE}(X_i) = \textnormal{argmax}_{W} \hat{Y}_i^{(W)}$. Then, we train a decision tree classifier to predict  $\Pi_{HTE}(X_i)$ given $X_i$. We create a personalized policy from this tree by assigning all individuals in the same terminal node (segment) to the majority treatment group of individuals in that segment. 

\subsubsection{Methods that learn directly from data} \label{sec:non_model_policy} 
The methods proposed in this section, \textit{\textbf{No-HTE}}, are useful when we are not able to train accurate HTE models to predict individual-level treatment effects, a scenario we sometimes see at Meta with extremely large datasets. These methods are inspired by prior efforts in the literature that segment feature space using conditional ATE \cite{dwivedi2020stable} or impurity \cite{kallus2017recursive}, and then generate policies directly. 

Concretely, for each segment $\mathcal{S}$, segment ATE, comparing treatment $k$ to control, can be computed by replacing $\mathcal{G}_k$ in Equation \eqref{eqn:ate} with $\mathcal{G}_k \cap \mathcal{S}$. Our goal is to search over different possible segments $\mathcal{S}$, by defining a splitting criterion $A(\mathcal{S}) = \max_{k=0,1,\ldots,K} A^{(k)} (\mathcal{S})$. This splitting criterion considers only the most responsive treatment in that segment. In practice, we implement this search using a tree with splitting criterion $A(\mathcal{S})$, splitting greedily or iteratively.

In the \textbf{greedy implementation} (Algorithm~\ref{algo:non_model_greedy} in the Appendix), a split is only considered if both the left and right child segments improve over the parent segment. In the \textbf{iterative implementation} (Algorithm~\ref{algo:non_model_iterative} in the Appendix), we run several iterations of splitting. In each iteration we keep the best segment, excluding it in the next run. A split is considered as long as one of the child segments improves the outcome compared to the parent node. One pitfall with this iterative implementation is that segments are not necessarily disjoint in feature space, so one individual could appear in several segments. We resolve this by always assigning the individual to the first found segment; other ways can be explored.

\textbf{Reject option.} For both implementations, if no segments are found because no eligible split exists, the policy defaults to assigning all individuals to the treatment group with highest ATE. In other words, this reduces to the standard randomized setting without personalization, and is in line with our interpretability goals of preferring simple policies if personalization does not bring benefits.
 
\subsubsection{Ensemble methods}
\label{sec:ensemble}
We can already generate interpretable tree-based policies using the methods described above. However, different policies may exhibit different strengths in different feature regions, and simply training trees with deeper depth does not necessarily improve the resulting policy. 
We leverage ensemble learning to identify such regions, with the hope of generating a better policy. Our motivation of combining different policies while remaining interpretable is shared by \cite{TuBasu2021}, but we do not use their method, because their method, being an intersection of segments from different trees, yields many more segments than suitable for our setting. While there exists other ways to ensemble policies, such as SuperLearner \citep{montoya2021optimal}, they result in non-interpretable policies. 

We ended up designing short trees, which we call guidance tree, that ensemble two policies with just one more feature split (see Figure \ref{fig:guidance_tree} in the Appendix for an example). \textit{\textbf{GUIDE-OPE}} finds this feature split using exhaustive search on the OPE criterion (Equation \eqref{eqn:ope}) to find one optimal feature split that becomes the top of the guidance tree. \textit{\textbf{GUIDE-ExploreExploit}} treats individual interpretable policies as ``arms", as in the contextual bandits literature \cite{slivkins2021introduction}, and uses explore-exploit to learn another policy on top of it that selects the candidate policy. See the Appendix for implementation details.

\section{Results on Publicly Available Data}
In this section, we evaluate the methods behind the interpretable personalized experimentation system on publicly available data. 

\subsection{Comparing Explanations of HTE Models}
\label{subsec: subgroups}

We compare the Distill-HTE method proposed in Section  \ref{sec:interpreter} against several other segment finding methods that take HTE model predictions as input: (1) Virtual Twins (VT) \citep{foster2011subgroup}; (2) R2P \citep{lee2020robust}, a recent, state-of-the-art method. We also compare to a black-box HTE model: a T-Learner that does not find segments but rather provides one prediction per individual. We train T-Learners using gradient boosted decision trees (GBDT) and decision tree (DT) base learners.

\textbf{Setup:} The COVID dataset was used in \cite{lee2020robust} and uses patient features as in an initial clinical trial for the Remdesivir drug, but generates synthetic outcomes where the drug reduces the time to improvement for patients with a shorter period of time between symptom onset to starting the trial. We chose this dataset because it has a specific finding of heterogeneity. The
second dataset, Synthetic A, was used in \cite{athey2016recursive}. %(we use the ``Design 1" setting). %but as noted in \cite{lee2020robust}, \note{possesses little homogeneity within subgroups} that is typical in reality. Still, we use it to compare against \cite{lee2020robust}. 
For R2P we used the implementation of R2P provided by the authors\footnote{https://github.com/vanderschaarlab/mlforhealthlabpub/tree/main/alg/r2p-hte}. For VT we used our own implementation, first training a black-box model (we use random forest (RF) and GBDT) to predict potential outcomes, then training a tree to explain the difference in potential outcomes.

\begin{table}
\scriptsize
\centering
\begin{tabular}{c l l l l l}
  \toprule 
  \multirow{2}{*}{\textbf{Data}} & \multirow{2}{*}{\textbf{Method}} & \multirow{2}{*}{PEHE} & Between-segment  &  Within-segment  \\ 
   &  &  & var &  var \\ 
  \midrule
   \multirow{6}{*}{\shortstack{COVID \\\cite{lee2020robust}}} & Virtual Twins RF & 0.94 $\pm$ 0.29 & 0.85 $\pm$ 0.31 &  2.88 $\pm$ 3.97 \\
   & Virtual Twins GBDT & 0.58 $\pm$ 0.20 & 0.07 $\pm$ 0.05 & 13.12 $\pm$ 1.20 \\
   & Distill-HTE & \textbf{0.20 $\pm$ 0.02} & 0.14 $\pm$ 0.05 & 11.03 $\pm$ 1.57 \\
   & R2P & 1.00 $\pm$ 0.21 & \textbf{1.00 $\pm$ 0.05} & \textbf{1.00 $\pm$ 1.03} \\
   & T-Learner GBDT & 0.58 $\pm$ 0.20 & -- & --\\
   & T-Learner DT & 1.46 $\pm$ 0.33 & -- & -- \\
   \midrule
   \multirow{6}{*}{\shortstack{Synthetic A\\ \cite{athey2016recursive}}} & Virtual Twins RF & 0.60 $\pm$ 0.07 & 0.24 $\pm$ 0.04 & 4.83 $\pm$ 0.64\\
   & Virtual Twins GBDT & \textbf{0.21 $\pm$ 0.02} & 0.05 $\pm$ 0.00 & 7.08 $\pm$ 0.75 \\
   & Distill-HTE & 0.26 $\pm$ 0.05 & 0.18 $\pm$ 0.04 & 6.69 $\pm$ 1.17 \\
   & R2P & 1.00 $\pm$ 0.28 & \textbf{1.00 $\pm$ 0.36} & \textbf{1.00 $\pm$ 1.97} \\
   & T-Learner GBDT & \textbf{0.21 $\pm$ 0.02} & -- & --\\
   & T-Learner DT & 0.86 $\pm$ 0.09 & -- & -- \\
   \bottomrule
\end{tabular}
\caption{Test-set performance of segment finding methods on publicly available datasets. For PEHE and within-segment variance, lower is better. For between-segment variance, higher is better. Best method for each column in bold.
Results are normalized w.r.t the state-of-the-art method R2P:
 a PEHE score of e.g. 0.2 ($20\%$) has 5 times less PEHE than R2P, while a score of 1.5 ($150\%$) has $50\%$ more.}
\label{table:pehe_variance}
\vspace{-0.55cm}
\end{table}

\textbf{Evaluation:} Since these datasets have synthetically injected ground truth treatment effects, we can measure the accuracy of each method. We use the Precision in Estimation of Heterogeneous Effect (PEHE) metric \citep{hill2011bayesian}:
$\sqrt{\frac{1}{N}  \sum_{i=1}^N \left((\hat{Y}_i^{(1)} - \hat{Y}_i^{(0)}) - (Y_i^{(1)} - Y_i^{(0)}) \right)^2}$.
Unlike bias and RMSE computed relative to ground truth treatment effects, PEHE requires accurate estimation of both counterfactual and factual outcomes \citep{johansson2016learning}. We also compute between- and within- segment variance. For each dataset we generate ten train-test splits, on which we compute the mean and standard deviation of estimates. 

\textbf{Hypothesis:} The best segmentation methods should have low PEHE and produce segments with low within-segment variance, and high between-segment variance. 

\textbf{Results:} Table \ref{table:pehe_variance} presents the results. We make a few observations: (1) \textit{Black-box vs. white-box}: As expected, GBDT T-Learners perform well as they do not have interpretability constraints, unlike all the other methods (VT, Distill-HTE, R2P), all of which modify standard decision trees while still remaining visualize-able as a tree. Yet, Distill-HTE tends to be far more accurate than R2P, in terms of PEHE. (2) \textit{Optimization criterion}: R2P, the only method of those presented here, that considers not only homogeneity within segments but also heterogeneity between segments, has the highest between-segment variance. Other methods that do not try to increase heterogeneity between segments do not fare so well on this metric. 
However, R2P does this at the expense of PEHE. (3) \textit{Impact of distillation}: While the T-Learner DT model did not perform well, being worst in terms of PEHE on all datasets, Virtual Twins RF, Virtual Twins GBDT and Distill-HTE that train modified decision trees have a marked improvement over T-Learner DT, suggesting that distilling a complex GBDT or RF teacher rather than learning a tree directly is beneficial, which agrees with the existing distillation literature. (4) \textit{HTE model class}: The choice of HTE model matters, with Virtual Twins not performing as well when using an RF HTE model compared to a GBDT HTE model. Similarly, GBDT T-Learners perform better than DT T-Learners. 

\textbf{Learnings:} Black-box HTE models are not always the most accurate on all datasets, and interpretable segmentation methods such as VT, Distill-HTE, and R2P can be as or more accurate on some datasets. We applied this lesson at the deployment phase of our system where black-box methods and interpretable methods are considered side-by-side.

\subsection{Comparing Policy Learning Methods}
We compare the methods described in Section \ref{sec: interpretable_policies} to (1) a black-box policy: training a T-Learner HTE model, then assigning each \textit{individual} to the treatment group with best predicted treatment effects; (2) a random policy: choosing the treatment for each unit uniformly at random. (3) On small datasets, we also compared to the PolicyTree \cite{zhou2018offline} method that uses exact tree search.

\textbf{Setup:} Besides the Synthetic A dataset described in \ref{subsec: subgroups}, we use other publicly-available datasets. The IHDP dataset \citep{hill2011bayesian}, commonly used in the causal inference literature (e.g \cite{johansson2016learning, sheth2021causebox} and many others), studied the impact of specialized home visits on infant cognition using mother and child features. A multiple-outcome dataset, Email Marketing \citep{Hillstrom} has 64k points, and visits, conversions, and money spent outcomes (see Appendix for details on how we constructed potential outcomes on this real dataset). We combine multiple outcomes as explained in Section \ref{sec:weights}.
The ensemble policies (GUIDE-ExploreExploit, GUIDE-OPE) are based on GreedyTreeSearch-HTE and Distill-Policy -- selected because of their individual performance. For the optimal tree search method, we used the implementation provided by the authors\footnote{https://cran.r-project.org/web/packages/policytree/index.html}.

\textbf{Evaluation:} Since these datasets have synthetically injected potential outcomes $\mathbf{Y}_i^{(k)}$, for each individual we know the optimal treatment group. We then evaluate different policies $\Pi$ using regret (against the optimal treatment): 
$\mathcal{R}(\Pi) = \sum_{i=1}^{n} \text{max}_k\mathbf{Y}_i^{(k)} - \mathbf{Y}_i^{(\Pi(X_i))}$
where $\Pi(X_i)$ is the treatment group $\in \{k=0, \ldots, K\}
$ prescribed by policy $\Pi$ for individual $i$. 

\textbf{Hypothesis:} The best policy generation methods should have low regret. Additionally, the more simple (less tree depth, and hence less segments) the policy, the easier it is to deploy, hence we study regret at a fixed level of complexity. 

\begin{table}[t!]
\scriptsize
\begin{tabular}{cll}
\toprule
\textbf{Data} & \textbf{Method} & \textbf{Test-Set Regret} \\
\midrule
 \multirow{10}{*}{\shortstack{IHDP\\ \cite{hill2011bayesian}}} & \textcolor{red}{Assign all to treatment 0} & 1.88 $\pm$ 0.17 \\
 & \textcolor{red}{Assign all to treatment 1} & \textbf{0.36 $\pm$ 0.06} \\
 & BlackBox & 1.00 $\pm$ 0.08 \\
 & \textcolor{blue}{OptimalTreeSearch \cite{zhou2018offline}} & 0.87 $\pm$ 0.09 \\
 & \textcolor{blue}{GreedyTreeSearch-HTE} & \textbf{0.77 $\pm$ 0.15} \\
 & \textcolor{blue}{Distill-Policy} & 1.00 $\pm$ 0.19 \\
 & \textcolor{blue}{No-HTE (Greedy and Iterative)*} & \textbf{0.36 $\pm$ 0.06} \\
 & \textcolor{green}{GUIDE-UniformExplore (guide depth=1)} & 1.05 $\pm$ 0.19 \\
 & \textcolor{green}{GUIDE-UniformExplore (guide depth=2)} & 0.96 $\pm$ 0.05 \\
 & \textcolor{green}{GUIDE-OPE (guide depth=1)} & 1.11 $\pm$ 0.16 \\
 \hline
\multirow{10}{*}{\shortstack{Synthetic A\\ \cite{athey2016recursive}}}& \textcolor{red}{Assign all to treatment 0} & 49.37 $\pm$ 1.77 \\
 & \textcolor{red}{Assign all to treatment 1} & 46.88 $\pm$ 1.73 \\
 & BlackBox & 1.00 $\pm$ 0.32 \\
  & \textcolor{blue}{OptimalTreeSearch \cite{zhou2018offline}} & 1.45 $\pm$ 0.28 \\
 & \textcolor{blue}{GreedyTreeSearch-HTE} & \textbf{0.03 $\pm$ 0.02} \\
 & \textcolor{blue}{Distill-Policy} & \textbf{0.05 $\pm$ 0.04} \\
 & \textcolor{blue}{No-HTE (Greedy and Iterative)*} & 46.88 $\pm$ 1.73 \\
 & \textcolor{green}{GUIDE-UniformExplore (guide depth=1)} & 0.01 $\pm$ 0.01 \\
 & \textcolor{green}{GUIDE-UniformExplore (guide depth=2)} & 0.02 $\pm$ 0.01 \\
 & \textcolor{green}{GUIDE-OPE (guide depth=1)} & \textbf{0.01 $\pm$ 0.01} \\
 \hline
  \multirow{10}{*}{\shortstack{Email Marketing\\ \cite{Hillstrom}}}
 & \textcolor{red}{Assign all to treatment 0} & 1.50 $\pm$ 0.06 \\
 & \textcolor{red}{Assign all to treatment 1} & 1.26 $\pm$ 0.10 \\
 & \textcolor{red}{Assign all to treatment 2} & 1.04 $\pm$ 0.02 \\
 & BlackBox & \textbf{1.00} $\pm$ \textbf{0.02} \\
 & \textcolor{blue}{GreedyTreeSearch-HTE} & 1.04 $\pm$ 0.02 \\
 & \textcolor{blue}{Distill-Policy} & 1.04 $\pm$ 0.02 \\
 & \textcolor{blue}{No-HTE (Greedy)} & 1.05 $\pm$ 0.03 \\
 & \textcolor{blue}{No-HTE (Iterative)} & 1.30 $\pm$ 0.05 \\
 & \textcolor{green}{GUIDE (UniformExplore and OPE)} & 1.04 $\pm$ 0.02 \\
\bottomrule
\end{tabular}

* denotes no segments were found, and the resulting policy assigned all individuals to one treatment group.

\caption{Regret (lower is better) of policy-generation methods on synthetic and semi-synthetic datasets. 
Methods colored in \textcolor{red}{red} do not personalize. Methods colored in black are black-box policies, \textcolor{blue}{blue} denotes interpretable policies, and \textcolor{green}{green} are ensembles that still remain interpretable.
Best method's regret in bold. If the best policy is no personalization, the next best policy is also bolded. If the best policy is an ensemble policy, the constituent policies are also bolded. Results are normalized w.r.t. the BlackBox regret: a regret score of e.g. 0.2 ($20\%$) has 5 times less regret than the BlackBox, while a regret of 1.5 ($150\%$) has $50\%$ more.}
\label{table:synthetic_results}
\vspace{-1.cm}
\end{table}

\textbf{Results:} Table \ref{table:synthetic_results} presents the results. 

\textit{Is personalization needed?} Interestingly, the majority of individuals in IHDP have positive treatment effects. Hence, there is limited benefit from personalization, where assigning all points to treatment 1 already yields the lowest relative regret of 0.36. Methods with a reject option (No-HTE-Greedy, No-HTE-Iterative) also performed well as they were able to pick up on this best policy and assign all points to a single treatment. 
 In contrast, Synthetic A exhibits some heterogeneity, with non-personalized policies (assigning all to treatment 0 or 1) having very large regret compared to BlackBox. Interpretable policies like GUIDE that ensembled GreedyTreeSearch-HTE and Distill-Policy achieved the lowest regret at 0.01, with the constituent policies themselves being not far off as well at 0.03 and 0.05 respectively. On this particular dataset, interpretable policies outperformed BlackBox policy. On Email Marketing, a real dataset, with more complicated heterogeneity, BlackBox performed the best.

\textit{Exact search.} We ran PolicyTree \cite{zhou2018offline} on IHDP and Synthetic A, two small datasets where the method did not face scalability issues. Despite performing exact search, the method does not necessarily have the smallest regret, as it is not an oracle and still has to estimate potential outcomes (only after which exact search happens).

\textbf{Learnings:} 
The performance of policy learning methods is dependent on the amount and type of heterogeneity in the dataset; no one policy
whether interpretable, black-box, or non-personalized, consistently performed well. Since we do not \textit{a priori} know how heterogeneous real datasets are, this impacted our system design, and we designed our policy generation stage to be highly modular (able to quickly add or remove different policy generation methods) as well as setup offline and online evaluation pipelines to rapidly test many different policies, thus allowing our system to adapt to the complexities of industry applications.

\subsection{Bridging Explanations and Policies}

\begin{figure}[t]
\centering
\includegraphics[width = 0.92\linewidth]{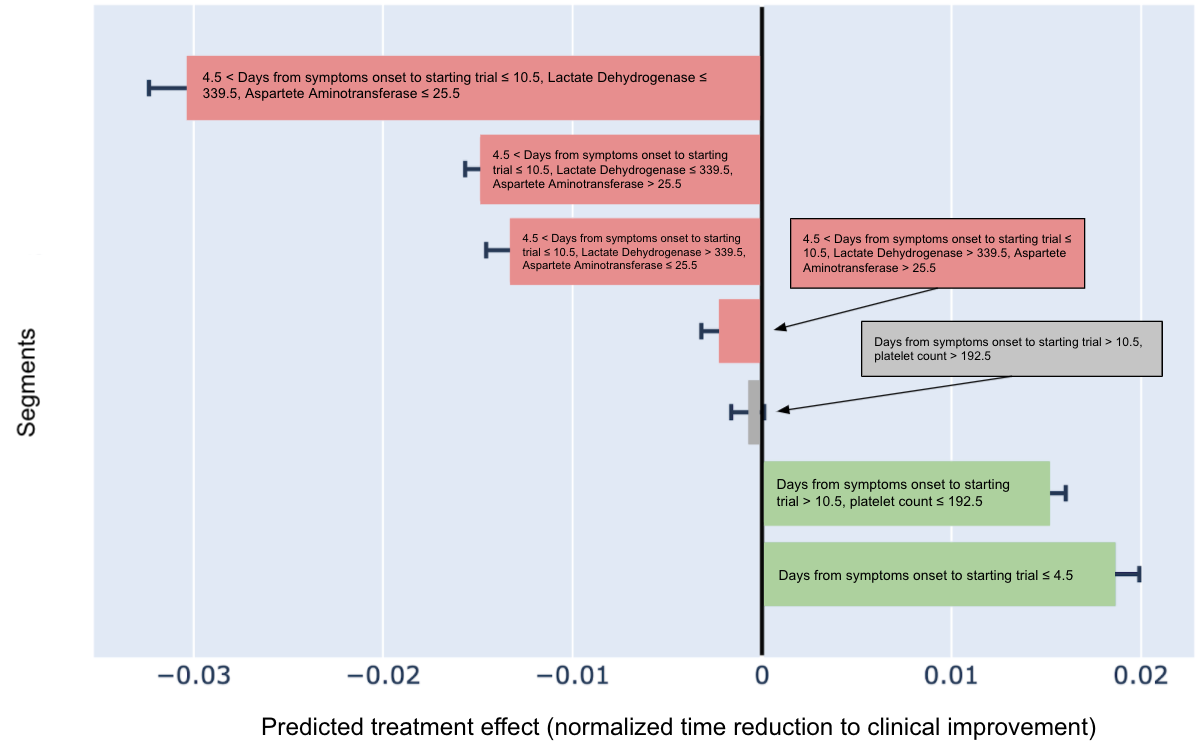}
\caption{Segments found by Distill-HTE method on COVID data. Best seen in digital format.}
\label{fig:covid_segments}
\end{figure}

\textbf{HTE model understanding:} Fig. \ref{fig:covid_segments} presents segments found by Distill-HTE on the COVID data. The segment with the most negative predicted treatment effect (first segment in red), at -0.03 +- 0.001, covers individuals who started taking  the drug between 4.5 and 10.5 days after the onset of symptoms and had Aspartete Aminotransferase and Lactate Dehydrogenase levels within normal ranges \citep{medicinenet}, suggesting that they were not extremely sick. It is unsurprising that they were not predicted to benefit as much from the drug.   

On the other hand, the segment with the most positive predicted treatment effect (last segment; green color) covers individuals who started taking the drug soon ($<=$ 4.5 days) after exhibiting COVID symptoms. These individuals were predicted to benefit the most from the drug, with treatment effect 0.018 +- 0.0006. This agrees with the finding that the Remdesivir drug results in a faster time to clinical improvement for the patients with shorter time from symptom onset to starting trial \citep{wang2020remdesivir}. 

\textbf{Policy generation:} A simple policy can be derived from the Distill-HTE segments, with all red segments assigned to not receiving the drug (as they are predicted to not benefit from it) and all green segments assigned to receiving the drug. This policy is different from the interpretable policies learned directly (Table \ref{table:synthetic_results}). For example, the GreedyTreeSearch-HTE policy is extremely simple: assigning to receive the drug only if days from symptoms onset to starting trial $<$ 11. Another example is No-HTE-Greedy and No-HTE-Iterative whose rather stringent splitting criteria did not find enough heterogeneity worth assigning different treatment groups to, and assigned all individuals to receive the drug. 

\section{Deployment at Meta}
\subsection{Personalizing UX Layout Using Interpretable Policies}
We tested the performance of the interpretable policy generation system on a UX design problem at Meta. With these new UX designs, individuals can see more media content per page. The hope is that this change will lead to more engagement. There are two UX designs (i.e. treatments) under consideration. There is one goal metric (larger is better) and one guardrail metric (must not decrease). The initial A/B test showed that Treatment 1 increases the goal metric on certain cohorts, and kept the guardrail metric neutral; Treatment 2 significantly increases the goal metric, but prohibitively decreases the guardrail metric. Our goal is to find a better tradeoff that preserves the goal metric gains seen in Treatment 2, but with the neutral guardrail metrics under Treatment 1.

To generate the interpretable policies, we trained black-box HTE models (X-learners) for each treatment, and applied our policy generation algorithms on them. 
These policies show that individuals who seldom view or engage with this type of media content should receive Treatment 1, while individuals with a history of consuming this content should receive Treatment 2.
Based on the OPE results (see Table~\ref{table:case_study_stories} for details), two interpretable policies were selected for online tests. With these particular metrics, even small improvements were of practical significance, and one interpretable policy was launched to the product. The launch realized more than 80\% of the organization's goal on the goal metric without adversely impacting the guardrail metric. 

\begin{table}[h]
\footnotesize
\begin{tabular}{cll}
\toprule
 \multirow{2}{*}{Method} & \multicolumn{1}{l}{\textbf{Test-Set OPE:}} & \multicolumn{1}{l}{\textbf{Test-Set OPE:}} \\
  & \multicolumn{1}{r}{\textbf{Goal Metric}} & \multicolumn{1}{r}{\textbf{Guardrail Metric}} \\
\midrule
  \textcolor{red}{Assign all to control} & 100.0 $\pm$ 0.7 & 100.0 $\pm$ 0.6 \\
  \textcolor{red}{Assign all to treatment 1} & 99.2 $\pm$ 0.7 & 100.0 $\pm$ 0.6 \\
  \textcolor{red}{Assign all to treatment 2} & 100.8 $\pm$ 0.7 & 99.8 $\pm$ 0.6 \\
  \textcolor{blue}{Distill-Policy} & \textbf{100.9 $\pm$ 0.7} & 99.9 $\pm$ 0.6 \\
  \textcolor{blue}{No-HTE (Greedy)} & 100.8 $\pm$ 0.7 & 100.0 $\pm$ 0.6 \\
\bottomrule
\end{tabular}
\caption{Offline policy evaluation (OPE) results on the UX design use case. The numbers are the estimated mean goal/guardrail metric values, larger is better. Non-personalized policies were colored in \textcolor{red}{red}, the two candidate personalized policies generated from our methods are in \textcolor{blue}{blue}. The policy that delivers the best goal metric improvement is bolded. Results are normalized w.r.t. the OPE of ``Assign all to control'' method.}
\label{table:case_study_stories}
\vspace{-0.55cm}
\end{table}

\subsection{Understanding Black-Box Personalization of Login Experience}
We used our HTE model understanding method on an account login flow problem at Meta. 
When a person is not automatically logged into their account after clicking on a notification, they could be directed to one of three experiences: (1) a traditional login screen, (2) a one-click login flow where a code is sent via email or text message to authenticate the user, or (3) a secondary login screen to allow the user to select between one-click login or a secondary identification flow. The product team wanted to increase login success while keeping guardrail metrics neutral.

The product team employed black-box HTE modeling to personalize this login experience. When deciding to ship the black-box personalized policy, the team used our Distill-HTE method to review heterogeneity present in the user base. The analysis revealed individuals with a recent password failure, who are using a device they don't own, or who haven't logged in recently respond most positively to the secondary login screen, and that the secondary login screen is the best treatment for individuals sharing a device. Users who are primarily active on the web browser site prefer the traditional login screen, while active app users respond best to the one-click login treatment. Furthermore, the analysis surfaced an issue in the team's experiment setup where a feature value was being incorrectly recorded. The team leveraged these findings to unblock shipping the personalized policy, fix the surfaced issue, and better understand the user base for future product changes.

\subsection{Product Team Interviews}
\label{sec: user_studies}
During development of this work, we conducted formal interviews with six potential internal customers from different product teams at Meta. The roles of the interviewees included: software engineer, data scientist, project manager, and marketer. With these backgrounds, interviewees exhibited a wide range of comfort interacting with and analyzing data. While data scientists were more comfortable with raw text output than other interviewees, all intervieews preferred small trees and the unrolled segments output shown in Figure \ref{fig:covid_segments} to raw text. Furthermore, those with less data analysis experience, such as marketers and project managers, preferred output like Figure \ref{fig:covid_segments} to small trees because interpreting trees was not immediately obvious to them. Overall, all potential customers believed the proposed interpretability analysis would benefit their understanding of HTE models and increase confidence in what the black-box models were doing for their product features at Meta. 
\vspace{-0.25cm}
\subsection{Deployment}
The Distill-HTE method is deployed within Meta, at scale on all experiments using black-box model personalization. Models for black-box personalized experiments are retrained recurrently, and the Distill-HTE method is reran after each retraining. We display the most recent analysis results, as shown in Figure \ref{fig:covid_segments}, within the experimentation system UI, thus allowing for an intuitive user experience. 
Additionally, we have deployed interpretable policies for multiple products at Meta. To deploy these policies, we use a modified version of the framework that deploys the black-box policies. Since interpretable policies are inherently simple to maintain interpretability, we can replace the black-box policy call with a few basic if-else statements. In practice, we implement the if-else blocks with comparison operators (less than, greater than, etc). Product teams frequently consider interpretable policies alongside black-box policies. 

The choice between black-box policies derived from HTE models and interpretable policies often depends on the ability of product teams utilizing such policies to maintain HTE models in production and the need to explain how exactly personalization is happening. While HTE models are resource and maintenance intensive, the ability to continuously retrain the model allows for adjustment to a dynamic user base. Conversely, interpretable policies are easy to implement and maintain, but may not perform best over the long-term without policy regeneration as the user base changes. Additionally, we have found that interpretable policies can be a better fit when app startup time is a constraint.

\vspace{-0.25cm}
\section{Conclusion}
The motivation for this work was three-fold. (1) HTE models sometimes overfit on extremely large datasets, and can be hard to interpret; (2) Interpretable policies can avoid some of the tech debt that black-box policies incur \citep{sculley2015hidden}. (3) Scalability constraints exclude optimal tree-based policy learning algorithms from deployment in production environments. Our two-stage system that understands black-box HTE models and generates interpretable personalized policies is deployed at Meta and runs on all personalized experiments. We hope that the practical experience and lessons shared in this paper can help other organizations wishing to incorporate interpretability techniques into their experimentation systems.

\vspace{0.25cm}
\noindent \textbf{Acknowledgements:} The authors would like to thank Rishav Rajendra, Zheng Yan, Chad Zhou, Sam Howie, Norm Zhou, and Ariel Evnine for valuable discussion and collaboration.

\bibliographystyle{ACM-Reference-Format}
\bibliography{main}

\clearpage
\section{Appendix}

\subsection{Algorithm Details}
In this section we provide more details of algorithms described in Section \ref{sec: interpretable_policies}. Define $\hat{M}_j(\mathcal{S}) = \sum_{i \in \mathcal{S}} \hat{Y}_i^{(j)}$ and $\hat{M}(\mathcal{S}) = \max_{j} \hat{M}_j(\mathcal{S})$. 
\begin{algorithm}
	\caption{GreedyTreeSearch-HTE}\label{algo:model_distill_prediction}
	\SetKwData{Left}{left}\SetKwData{This}{this}\SetKwData{Up}{up}
	\SetKwFunction{Union}{Union}\SetKwFunction{FindCompress}{FindCompress}
	\SetKwInOut{Input}{Input}\SetKwInOut{Output}{Output}
	\Input{Maximal tree depth $d$, dataset $\mathcal{D}$, predictions from HTE model  $\lbrace (\hat{Y}_i^{(0)}, \ldots., \hat{Y}_i^{(K)}) \rbrace_{i=1}^N$}
	\Output{A list of segments: $segments$.}
	1. Set $depth:=0$ and $segments = \lbrace \mathcal{D} \rbrace$.
	
	2. Add segments to $segments$ using greedy search:
	
	\While {$depth \leq m$} {
	Set $nodes:= \lbrace \rbrace $ \\
    \For  {$\mathcal{S} \in segments$}   {
			Split $\mathcal{S}$ into $\mathcal{S}^*_l$ and $\mathcal{S}^*_r$ such that:  
            \begin{align*}
			   & (\mathcal{S}^*_l, \mathcal{S}^*_r) = \text{argmax}_{\hat{M}(\mathcal{S}_l) +  \hat{M}(\mathcal{S}_r)  >  \hat{M}(\mathcal{S})}  (\hat{M}(\mathcal{S}_l) +  \hat{M}(\mathcal{S}_r)).
			\end{align*} 
			If such $\mathcal{S}^*_l$ and $\mathcal{S}^*_r$ exist, we add them to $nodes$; otherwise we add $\mathcal{S}$ to $nodes$.
	} 
	Let $segments := nodes$ and $depth := depth+1$.
	}
\end{algorithm}

\begin{algorithm}[h!]
	\caption{No-HTE-Greedy}\label{algo:non_model_greedy}
	\SetKwData{Left}{left}\SetKwData{This}{this}\SetKwData{Up}{up}
	\SetKwFunction{Union}{Union}\SetKwFunction{FindCompress}{FindCompress}
	\SetKwInOut{Input}{Input}\SetKwInOut{Output}{Output}
	\Input{Maximal tree depth $d$, dataset $\mathcal{D}$.}
	\Output{$\text{segments}$}
	1. Set $\text{depth}=0$, $\text{segments} = \lbrace \mathcal{D} \rbrace$.
	 
	2. Split the segments:  

	\While {$\textnormal{depth} \leq d$} {
	Set $\text{nodes}:= \lbrace \rbrace $ \\
    \For  {$\mathcal{S} \in \textnormal{segments}$}   {
			Consider splits that satisfy $\min (A(\mathcal{S}_l), A(\mathcal{S}_r)) > A(\mathcal{S})$, find the optimal split defined as follows:  
            \begin{align*}
			   & (\mathcal{S}_l^*, \mathcal{S}_r^*) = \text{argmax}_{\mathcal{S}_l, \mathcal{S}_r} \min (A(\mathcal{S}_l), A(\mathcal{S}_r)). \label{eq:algo_1}
			\end{align*}  \\
			If such $\mathcal{S}_l^*$ and $\mathcal{S}_r^*$ exist, add them to $\text{nodes}$; otherwise add $\mathcal{S}$ to $\text{nodes}$.
	} 
	Let $\text{segments} := \text{nodes}$ and $\text{depth} := \text{depth}+1$.
	}
\end{algorithm}

\begin{algorithm}
	\caption{No-HTE-Iterative}\label{algo:non_model_iterative}
	\SetKwData{Left}{left}\SetKwData{This}{this}\SetKwData{Up}{up}
	\SetKwFunction{Union}{Union}\SetKwFunction{FindCompress}{FindCompress}
	\SetKwInOut{Input}{Input}\SetKwInOut{Output}{Output}
	\Input{Maximal tree depth $d$, number of iterations $t$, dataset $\mathcal{D}$.}
	\Output{$\text{segments}$}
	1. Set $\text{iterations}:=0$ and $\text{segments} = \lbrace \rbrace$.
	
	2. Add to $\text{segments}$ iteratively
	
	\While {$\textnormal{iterations} \leq t$} {
	    Step 1 to 2 in Algorithm~\ref{algo:non_model_greedy}, but change all $\min$ to $\max$. Denote the resulting list of segments as $C$.
	   
	    Add to $\text{segments}$ the $\mathcal{S}$ in $C$ that maximizes $A(\mathcal{S})$ and remove the data points in $\mathcal{S}$ from $\mathcal{D}$.
	    
	    Set $\text{iterations} := \text{iterations} + 1.$
	}
\end{algorithm}

\subsection{Ensemble Algorithm Details}
In this section we provide more details of algorithms described in Section \ref{sec:ensemble}. Suppose we have access to $Q$ policies $\Pi_1, \Pi_2, \ldots, \Pi_Q$. We wish to train an ensemble policy $\widetilde{\Pi}$ that uses all or a subset of the policies $\Pi_1, \Pi_2, \ldots, \Pi_Q$ while still remaining interpretable. For ease of notation, we assume the trained policies $\Pi_1,\ldots,\Pi_Q$ were obtained from another split of the dataset and we can safely use $\mathcal{D}$ as the validation set on which we learn the ensemble policy.

\begin{algorithm}
	\caption{GUIDE-ExploreExploit}\label{algo:ensemble_prediction}
	\SetKwData{Left}{left}\SetKwData{This}{this}\SetKwData{Up}{up}
	\SetKwFunction{Union}{Union}\SetKwFunction{FindCompress}{FindCompress}
	\SetKwInOut{Input}{Input}\SetKwInOut{Output}{Output}
	\Input{Maximal tree depth $d$, dataset $\mathcal{D}$, predictions $\lbrace (\hat{Y}_i^{(0)}, \ldots., \hat{Y}_i^{(K)}) \rbrace_{i=1}^N$, $Q$ trained interpretable policies $\Pi_1,\ldots,\Pi_Q$.} 
	\Output{A list of segments: $segments$.}
	1. Generate a dataset with randomly selected policies on individuals
	
	\For {$i=1$ \KwTo $N$}{
	  Randomly select a policy from $\Pi_1,\ldots,\Pi_Q$. Let $A_i$ be the index of this policy, $A_i \in {1, \ldots, Q}$. 
	  Apply this policy to individual $i$, assume the policy assigns treatment group $k \in \{0,1,\ldots,K\}$. Initialize $O_i = [0,\ldots,0] \in \mathbb{R}^{K + 1}$. Let $W_i$ be the treatment received by individual $i$ in the dataset.\\
	  If $k==W_i$, set $O_i^{(k)} := Y_i$. Otherwise, set $O_i^{(k)} :=\hat{Y}_i^{(k)}$.
	}
	We now have a new dataset of the form $(X_i, A_i \in \{1,2,\ldots,Q\}, O_i)$.
	
	2. Apply Algorithm~\ref{algo:model_distill_prediction} to the new dataset, using $(O_i^{(0)}, \ldots, O_i^{(K)})$ in place of $(\hat{Y}_i^{(0)}, \ldots, \hat{Y}_i^{(K)})$.
\end{algorithm}

\textit{\textbf{GUIDE-ExploreExploit}} is inspired by the explore-exploit paradigm in the contextual bandits literature \cite{slivkins2021introduction}. To perform this offline, we use HTE outcome predictions when the observed outcome is not available in the dataset. See Algorithm~\ref{algo:ensemble_prediction} for the implementation. The ensemble policy $\tilde{\Pi}$ is generated using Algorithm \ref{algo:model_distill_prediction} but with the HTE predictions $(\hat{Y}_i^{(0)}, \ldots, \hat{Y}_i^{(K)})$ replaced by $(O_i^{(0)}, \ldots, O_i^{(K)})$.

% \begin{algorithm}
% 	\caption{Off-Policy Evaluation (OPE) with Inverse Propensity Scoring}\label{algo:ope_ips}
% 	\SetKwData{Left}{left}\SetKwData{This}{this}\SetKwData{Up}{up}
% 	\SetKwFunction{Union}{Union}\SetKwFunction{FindCompress}{FindCompress}
% 	\SetKwInOut{Input}{Input}\SetKwInOut{Output}{Output}
% 	\Input{Policy $\Pi$, dataset $\mathcal{D}$, propensity score $\lbrace p_i  \rbrace_{i=1}^N$ for all individuals ($p_i = \frac{1}{K+1}$ for randomized data with $K+1$ treatment groups).} 
	
% 	\Output{The estimated policy value.}
% 	1. Generate the index set where policy decision matches the observed treatment decision in $\mathcal{D}$: $\mathcal{I} = \lbrace i | W_i == \Pi(X_i) \rbrace.$
	
% 	2. Output $\frac{1}{N}\sum_{i \in \mathcal{I}} Y_i / p_i(W_i)$.
% \end{algorithm}
\textit{\textbf{GUIDE-OPE}} generates a policy ensemble by maximizing off-policy evaluation (Equation \eqref{eqn:ope}) at each split. Here, we aim to find one feature split such that the left and right children uses a different candidate policy. 
Concretely, suppose we split dataset $\mathcal{D}$ into $\mathcal{D}_l$ and $\mathcal{D}_r$, let $\tilde{\Pi}$ be the ensemble policy that applies $\Pi_k$ to $\mathcal{D}_l$ and $\Pi_q$ to $\mathcal{D}_r$. 
To create $\tilde{\Pi}$, we use exhaustive search to find the optimal feature split and candidate policies $(\mathcal{D}_l^*, \mathcal{D}_r^*, \Pi_k^*, \Pi_q^*)$ that solves $ \max_{\mathcal{D}_l, \mathcal{D}_r} \max_{1 \leq k \neq q \leq Q} \text{OPE}(\mathcal{D}, \tilde{\Pi})$. We assign to individual $i$ policy $\Pi_k^*$ if individual $i \in \mathcal{D}_l^*$ and $\Pi_q^*$ otherwise.

\begin{figure}
\centering
\includegraphics[width=0.74\linewidth]{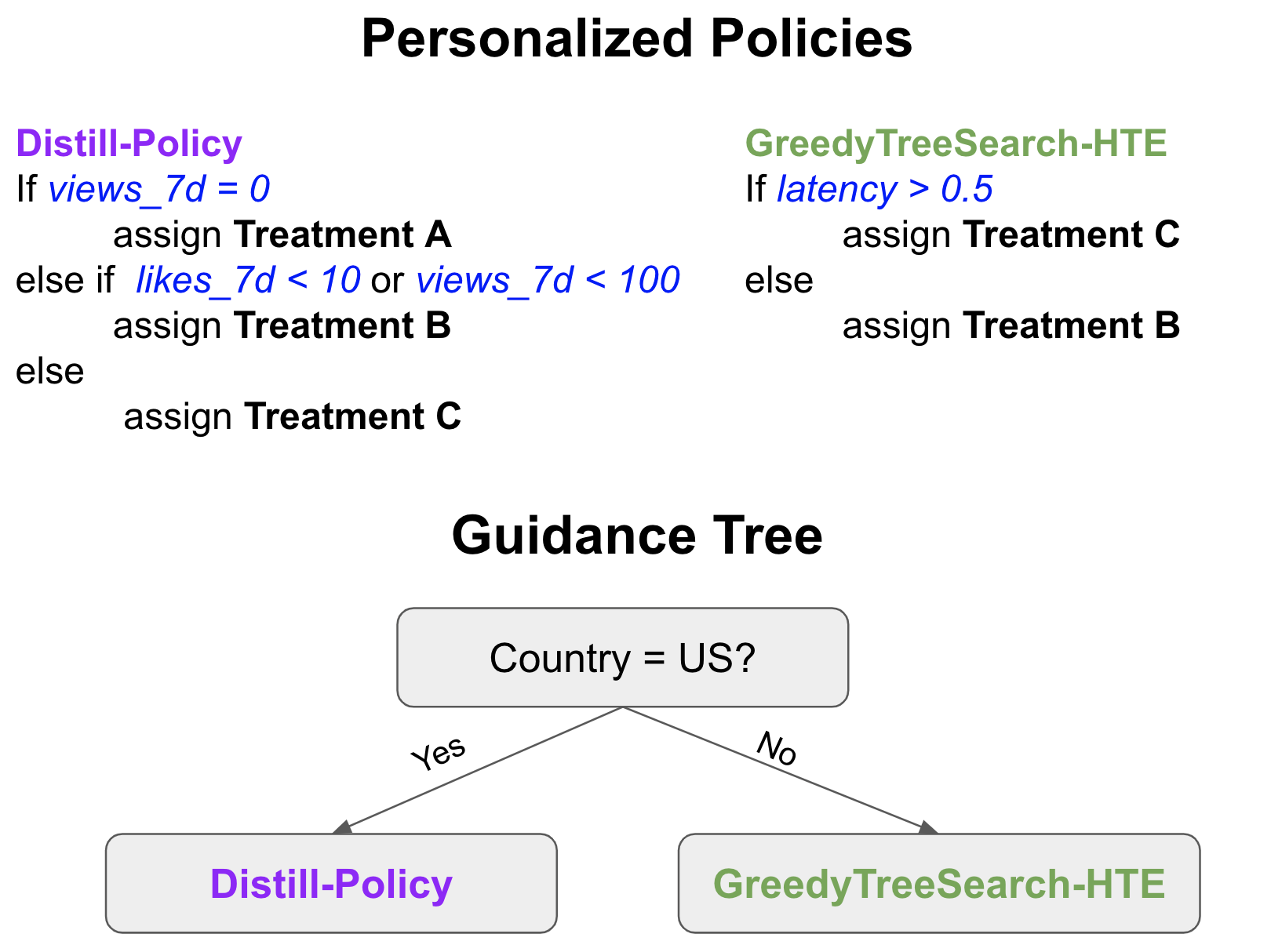}
\caption{\textit{Top:} Personalized policies. \textit{Bottom:} Guidance tree learned using GUIDE-ExploreExploit or GUIDE-OPE, to ensemble two interpretable policies while remaining interpretable.}
\label{fig:guidance_tree}
%\vspace{-0.55cm}
\end{figure}

\subsection{Data Generation Details}
The Email Marketing dataset \citep{Hillstrom}, first introduced in the public MineThatData\footnote{\url{https://blog.minethatdata.com/2008/03/minethatdata-e-mail-analytics-and-data.html}} Data Mining challenge, is a real dataset from an experiment where customers were randomized into receiving one of three treatments. To generate potential outcomes, for each individual $i$ in treatment group $k$, we searched for the 5-nearest-neighbors in treatment group $k'$, averaging their outcomes to get the potential outcome for individual $i$ for treatment group $k'$. To compute distance between individuals, we used Euclidean distance in feature space.

\subsection{Training Details}
Unless otherwise mentioned, the HTE models we train are T-learners consisting of GBDT base learners. In general, we use 40\% of the data as the test set on which we report results, and 30\% of the remaining 60\% as the validation set. We learn individual policies on the training set. When learning ensemble policies, we learn the ensemble on the validation set. 

\end{document}